\documentclass[11pt,a4paper]{article}
\usepackage[hyperref]{acl2021}
\usepackage{times}
\usepackage{latexsym}
\usepackage{CJKutf8}
\usepackage{graphicx}
\usepackage{booktabs}
\usepackage{multirow}
\usepackage{amsmath}
\usepackage{bm}
\usepackage{url}
\usepackage{tabularx}
\usepackage{verbatim}
\usepackage{caption}
\usepackage{amsfonts}
\usepackage{subfig}

\newcommand{\tabincell}[2]{\begin{tabular}{@{}#1@{}}#2\end{tabular}}
\usepackage{xcolor}
\usepackage{soul}
\usepackage{float}

\definecolor{word}{RGB}{255, 255, 255} 
\definecolor{support}{RGB}{140,98,175}
\definecolor{information}{RGB}{91, 155, 213} 
\definecolor{guidance}{RGB}{218,165,32}
\definecolor{restatement}{RGB}{0, 176, 80}
\definecolor{interpretation}{RGB}{192, 0, 0}

\newcommand{\supportstrategy}[1]{
  \begingroup
  \sethlcolor{support}
  \textcolor{word}{\hl{#1}}
  \endgroup
}

\newcommand{\informationstrategy}[1]{
  \begingroup
  \sethlcolor{information}
  \textcolor{word}{\hl{#1}}
  \endgroup
}

\newcommand{\interpretationstrategy}[1]{
  \begingroup
  \sethlcolor{interpretation}
  \textcolor{word}{\hl{#1}}
  \endgroup
}

\newcommand{\restatementstrategy}[1]{
  \begingroup
  \sethlcolor{restatement}
  \textcolor{word}{\hl{#1}}
  \endgroup
}

\newcommand{\guidancestrategy}[1]{
  \begingroup
  \sethlcolor{guidance}
  \textcolor{word}{\hl{#1}}
  \endgroup
}

\usepackage{microtype}

\aclfinalcopy 

\title{PsyQA: A Chinese Dataset for Generating Long Counseling Text for Mental Health Support}

\author{Hao Sun$^1$\thanks{\ \ Equal contribution.} , Zhenru Lin$^2$\footnotemark[1] , Chujie Zheng$^1$, Siyang Liu$^3$, Minlie Huang$^1$\thanks{\ \ Corresponding author.} \\
  \small $^1$The CoAI group, DCST, Institute for Artificial Intelligence, State Key Lab of Intelligent Technology and Systems, \\
  \small $^1$Beijing National Research Center for Information Science and Technology, Tsinghua University, Beijing 100084, China \\
  \small $^2$Dept. of Electronic Engineering, Tsinghua University, Beijing 100084, China \\
  \small $^3$Tsinghua-Berkeley Shenzhen Institute, Tsinghua Shenzhen International Graduate School, \\ 
  \small $^3$Tsinghua University, Shenzhen, China \\
  {\small \tt \{h-sun20,linzr18\}@mails.tsinghua.edu.cn, aihuang@tsinghua.edu.cn} \\
}

\date{}

\begin{document}

\begin{CJK*}{UTF8}{gbsn}
\maketitle
\begin{abstract}
  Great research interests have been attracted to devise AI services that are able to provide mental health support.
  However, the lack of corpora is a main obstacle to this research, particularly in Chinese language.
  In this paper, we propose PsyQA, a Chinese dataset of psychological health support in the form of question and answer pair. 
  PsyQA is crawled from a Chinese mental health service platform, and contains 22K questions and 56K long and well-structured answers.
  Based on the psychological counseling theories, we annotate a portion of answer texts with typical strategies for providing support, and further present in-depth analysis of both lexical features and strategy patterns in the counseling answers.
  We also evaluate the performance of generating counseling answers with the generative pretrained models. 
  Results show that utilizing strategies enhances the fluency and helpfulness of generated answers, but there is still a large space for future research.


\end{abstract}

\section{Introduction}
The burden of mental disorders continues to grow with significant impacts on human health and social development \cite{world2011global,WHO2020}. 
As an effective therapy for mental disorders \cite{reynolds2013impact}, online mental health counseling, which mostly refers to communicating anonymously, has become popular in recent years \cite{fu2020effectiveness}.


Great research interests have been endeavored to devise AI services that are able to provide mental health support \cite{bucci2019digital, liu-etal-2021-towards}.
Based on the online-text psychotherapy corpora, previous works have utilized text mining techniques to detect empathy \cite{sharma2020computational, zheng-etal-2021-comae}, linguistic development of counselors \cite{zhang2019finding}, and self‐injurious thoughts and behaviors \cite{franz2020using}. 
However, the research of text-based mental health counseling is still largely limited due to the lack of relevant corpora, particularly in Chinese language.

\begin{figure}[tbp]
    \centering
    \includegraphics[width=\linewidth]{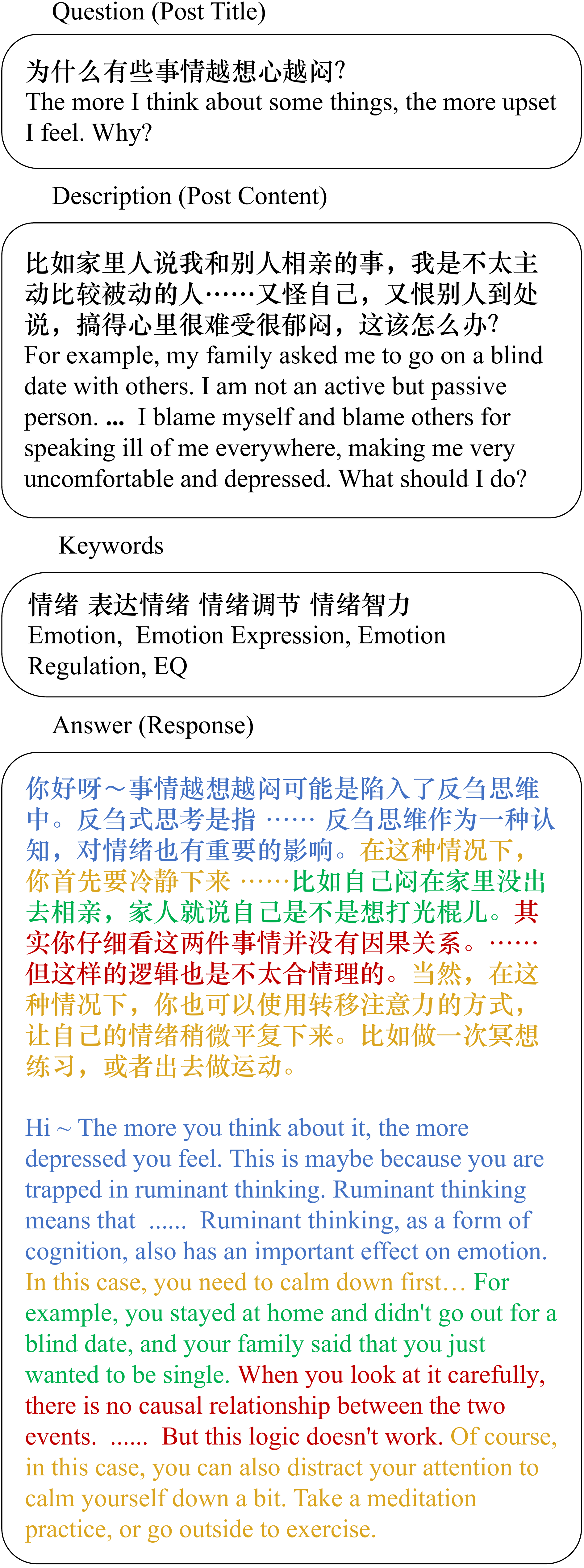}
    \caption{An example from PsyQA. $(Question, Description, Keyword)$ triples are posted by help-seekers while $Answer$ is provided by help-supporters. Different strategies in the answer are colored differently. Strategies \informationstrategy{Information}, \interpretationstrategy{Interpretation}, \restatementstrategy{Restatement}, and \guidancestrategy{Direct Guidance} are used in this answer. Note that a question may have multiple answers.}
    \label{fig:dataset_exam}
\end{figure}

To this end, we collect \textbf{PsyQA} in this work, a Chinese dataset of \textbf{Psy}chological health support in the form of \textbf{Q}uestion-\textbf{A}nswer pair.
An example data of PsyQA is shown in Figure \ref{fig:dataset_exam}.
In each example, the \textbf{\textit{question}} along with a detailed \textbf{\textit{description}} and several \textbf{\textit{keyword}} tags is posted by an anonymous help-seeker, where the description generally contains dense persona and emotion information about the help-seeker. 
The \textbf{\textit{answer}} is usually quite long (524 words on average).
The answers are replied asynchronously from the well-trained volunteers or professional counselors, and contain both the detailed analysis of the seeker's problem and the guidance for the seeker. 
Moreover, a portion of the answers are also additionally annotated by professional workers with typical support \textbf{\textit{strategies}}, which are based on the psychological counseling theories \cite{hill2009helping}.

Our collected PsyQA has three distinct characteristics. 
Firstly, the corpus covers abundant mental health topics from 9 categories including emotion, relationships, and so on 
(refer to Appendix for topic statistics).
Secondly, the answers in PsyQA are mostly provided by experienced and well-trained volunteers or professional counselors.
Thirdly, we provide support strategy annotations for a portion of answers, which can greatly facilitate future research on our corpus.
As will be shown later, there are not only lexical features in the texts of different support strategies (Section \ref{sec:feature}), but also explicit patterns of strategy organization and utilization in the answers (Section \ref{sec:strategy_seq}).



To validate whether existing models can generate long counseling answers to mental health questions, we conduct experiments on both strategy identification (Section \ref{sec:strategy_clf}) and answer generation (Section \ref{sec:ans_gen}).
We find that the contextual information greatly benefits the performance of support strategy identification.
Experimental results also demonstrate that utilizing support strategies improves the answers generated by the models in terms of their language fluency, coherence, and the ability to be on-topic and helpful. 
However, there is still much room for further research compared to the answers written by well-trained volunteers or professional counselors.


Our contributions are summarized as follows:
\begin{itemize}
    \item We collect PsyQA, a high-quality Chinese dataset of psychological health support in the form of QA pair. 
    The answers in PsyQA are usually long, which are provided by well-trained volunteers or professional counselors.
    
    \item We annotate a portion of answer texts with a set of strategies for mental health support based on psychological counseling theories.
    Our analysis reveals that there are not only typical lexical features in the texts of different strategies, but also explicit patterns of strategy organization and utilization in the answers.
    
    \item We conduct experiments of both strategy identification and answer text generation on PsyQA. 
    Results demonstrate the importance of using support strategies, meanwhile indicating a large space for future research.
\end{itemize}

\section{Related Work}
Our work primarily concerns linguistic behavior for counseling, NLP for mental health detection and therapy, and text-based mental health-related datasets.

\subsection{Linguistic Behaviors in Counseling}
Hill's model \cite{hill2009helping} consists of three stages: exploration, insight, and action in which helpers guide clients in exploring their thoughts and feelings, discovering the origins and consequences of maladaptive thoughts and behaviors, and acting on those discoveries to create positive long-term change. We draw on Hill's model and apply it to formulate the answer in the PsyQA dataset. 

Some previous work explored how mental health support is sought and provided.
For example, some studies measure how the language of comments in Reddit mental health communities influences risk to suicidal ideation in the future \cite{de2017language}, and seek to understand how counselors' behaviors develop over time \cite{zhang2019finding}.
While these previous studies model implicit linguistic behaviors of counselors, we focus on linguistic strategy development in a long psychological response, considering the strategies as a skeleton to generate the whole response.

\subsection{NLP for Mental Health Detection and Therapy}

Some prior work analyzed the posts and blogs of users with the rise of social networking sites (SNS), attempting to employ NLP techniques to detect depression \cite{tadesse2019detection, yates2017depression}, suicidal ideation \cite{zirikly2019clpsych, cao2019latent}, and other general mental health problems \cite{xu2020inferring}.
In another line of work, some researchers endeavored to construct ``therapybots'' \cite{fitzpatrick2017delivering, inkster2018empathy}, and focused on therapy and attempted to create dialogue agents to provide therapeutic benefit, where the effectiveness of web-based cognitive-behavioral therapeutic (CBT) apps or mobile mental well-being apps are explored. 
Adopting a more straightforward method, we make the machine generate answers to a detailed question, mimicking a mental health counselor. 
Though the ultimate goal is to develop systems for real-world treatment, there is still a long way to go in this direction and our corpus can be the first step towards building intelligent systems for this purpose, and offers the opportunity for studying the effectiveness of using explicit strategies in the systems.

\begin{table*}[t]
    \centering
    \includegraphics[width=\linewidth]{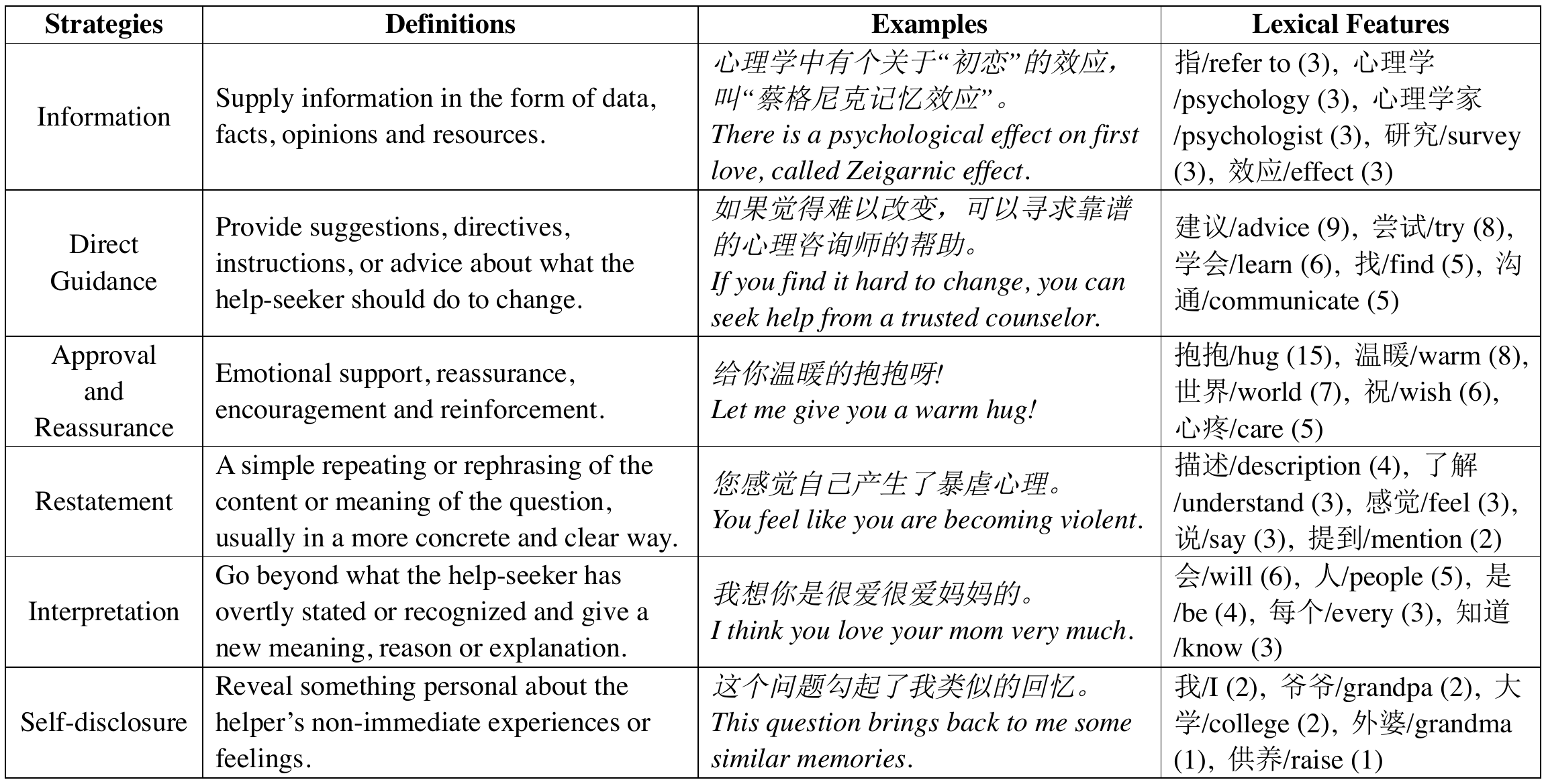}
    \caption{
        The definition and example of different strategies in our guideline, together with the lexical features of the strategies in our annotated dataset.  The rightmost column displays the top 5 words associated with each strategy. The rounded z-scored log odds ratios are in the parentheses. A word may appear in multiple parts of speech. For example, "warm" in Chinese can be either an adjective or a verb.
    }
    \label{tab:definition}
    \vspace{-2mm}
\end{table*}

\subsection{Text-based Mental Health-Related Datasets}
There are some datasets for mental health detection and therapy.
However, most of them are collected from general social networking sites such as Twitter, Reddit, and Weibo \cite{harrigian2020state}. 
General social networking sites contain irrelevant posts or unprofessional responses, which might put NLP systems trained on these corpora at huge risk.
Thus, some previous work focused on the counseling part in the online mental health communities (forums), such as TeenHelp \cite{franz2020using}, TalkLife \cite{sharma2020computational}.
In Chinese domain, \citet{efaqa-corpus-zh:petpsychology} collected a public counseling conversation dataset by crawling. However, most responses in this dataset are short and general without any suggestion.
Crisis Text Line \cite{althoff2016counseling} presents the best mental health counseling dataset up to now. It contains a large-scale multi-turn counseling conversation by experienced volunteer counselors\footnote{Unfortunately, there is no public access to this dataset.}.
Different from Crisis Text Line, PsyQA focuses on Chinese long-text response in a single-turn asynchronous counseling conversation.

From the perspective of the mental health domains, most of the prior work is focusing on single-domain like depression, suicidal ideation, and eating disorders \cite{harrigian2020state}. Instead, PsyQA contains all sorts of general mental health disorders, concerning nine topics labeled by help-seekers including self-growth, emotion, love problem, relationships, behaviors, family, treatment, marriage, and career.

\section{Data Collection}

\subsection{Data Source}

Our dataset is crawled from the Q\&A column of Yixinli (\url{xinli001.com/qa}). 
Yixinli is a Chinese mental health service platform with about 22 million users and over six hundred professional counselors.
In its Q\&A column, anonymous users post questions about their daily-life worries, and well-trained volunteers or professional counselors answer them with detailed analysis and guidance in the form of organized long texts.
More than 0.25 million Q\&A pairs are on this platform, with abundant topics ranging from personal development and relationships, to mental illnesses. 
Yixinli manually review and block unsafe contents
To avoid potential ethical risks and ensure the quality of the data. 
We calculate that in our dataset, the help-supporters have ever answered over 250 questions on average. Besides, 8\% answers are from help-supporters who are State-Certificated Class 2 Psychological Counselors, and 35\% answers are from volunteers hired by Yixinli.

\subsection{Data Cleaning}

We removed personal information, duplicate line breaks, emojis, website links and advertisements by rule-based filtering.
Besides, to ensure a higher quality, only those answers with more than 100 words were retained. 
It is inevitable that there exist some unrelated posts in raw websites. 
To remove such posts, we tried to filter out questions that are not actually seeking for mental health support based on keywords (topics) given by the poster, such as the questions that ask about the meaning of a psychological term (keyword: popular science) or discuss the latest news (keyword: hot news).

\subsection{Strategy Annotation}

We analyzed multiple high-quality answers in our corpus and found that the strategies employed by the help-supporters are consistent with
Helping Skills System (HSS) \cite{hill2009helping}. 
Moreover, we observed that the strategy sequence patterns are similar to some degree. 
Thus, we assumed that a whole answer is realized through an organized strategy sequence, which may reveal the common layout of high-quality responses from mental health counselors. 
To facilitate further research on strategies in text-based mental health support, we then present the process that we annotated the answers with span-level strategies.

\citet{hill2009helping} provides a taxonomy of language helping skills or strategies for mental health counselors. 
We chose a subset of strategies according to the general online counseling situation, while also corresponds to the guideline for help-supporters from Yixinli web.
Table \ref{tab:definition} shows the list of our chosen strategies with their definitions and examples.

We randomly sampled 4,012 questions (about 17.9\%) in our dataset and picked their highest-voted answers (similar to Quora, \url{quora.com}). 
Then we recruited and trained 9 workers to annotate the answers following our guideline.\footnote{
All annotators in this work are compensated for 60 in CNY per hour, which is reasonable compared with the mean income of urban residents in China.}
We leveraged Doccano\footnote{\url{https://github.com/doccano/doccano}}, an open-source text annotation tool, for the workers to annotate the text. 
In each task, the workers were shown a Q\&A pair and asked to label one or more consecutive sentences (a text span) with a strategy.
The workers were allowed to ignore the sentences that did not match the definition of any strategy, which would be automatically labeled as \textit{Others}.

\subsection{Annotation Quality Control}

The workers were required to read the guideline and the provided annotated examples before annotation. 
To verify the effectiveness of training, we asked them to annotate 100 examples before formal annotation, which were revised by psychology professionals for feedback. 
We repeated the above process until the workers were able to annotate the cases almost correctly. 
After annotation, to check the quality of labels, we randomly sampled 200 annotated Q\&A pairs, gave them to 2 examiners (both are graduate students of Clinical Psychology) to pick out incorrect labels, and calculated the consistency proportion.
Results are shown in Table \ref{tab:stra_consis}.
More than 98\% of the strategy labels are consistent with at least one examiner, indicating the reliability of strategy annotation.

\begin{table}[t]
    \centering
    \scalebox{0.8}{
        \begin{tabular}{cccc}
            \toprule
                \textbf{Strategy $\bm{\backslash}$ Consis.} & \textbf{1/2} & \textbf{2/2} & \textbf{\# Samples} \\
            \midrule
                Restatement & 0.981 & 0.932 & 162 \\
                Appro.\& Reass. & 0.994 & 0.982 & 165 \\
                Interpretation & 0.961 & 0.820 & 610 \\
                Information & 0.990 & 0.912 & 102 \\
                Self-disclosure & 1.000 & 0.932 & 162 \\
                Direct Guidance & 0.992 & 0.870 & 509 \\
            \midrule
                Overall & 0.980 & 0.876 & 1,616 \\
            \bottomrule
        \end{tabular}
    }
    \caption{Consistency proportion of strategy annotation samples. 
    1/2 means consistency with at least one examiner, and 2/2 means consistency with both examiners. }
    \label{tab:stra_consis}
\end{table}

\section{Corpus Analysis}

\subsection{Statistics}
\label{sec:stats}

Table \ref{tab:sum_sta} shows the statistics of our dataset. 
The long answer text is a distinct feature of our dataset, and the annotated answers are even longer. 
There is also a wide variety of strategies in the answers (6.66 ones and 3.65 distinct ones per answer), and we will further analyze the patterns of strategy utilization in Section \ref{sec:strategy_seq}.

Note that our dataset covers 9 broad topics (e.g. \textit{self-growth}, \textit{emotion}, etc.) and a wide range of subtopics (e.g.\textit{personality improvement}, \textit{emotion regulation}, etc.)\footnote{
Please refer to Appendix \ref{sec:app_que_key} for the categories of topics and subtopics, together with detailed statistics of topics.
}, from which the seekers can choose as the question keywords.

\begin{table}[t]
    \centering
    \scalebox{0.8}{
        \begin{tabular}{p{15em}c}
            \toprule
            \textbf{Criteria} & \textbf{Statistics} \\ 
            \midrule
            \# Questions     & 22,346     \\ 
            \# Answers   & 56,063  \\ 
            \# Characters per question  & 21.6   \\ 
            \# Characters per description   & 168.9 \\ 
            \# Characters per answer  & 524.6 \\ 
            \cmidrule{1-2}   
            \tabincell{l}{\# Annotated questions / answers}   & \tabincell{c}{4,012 / 4,012\\ (17.9\% / 7.1\%)} \\ 
            \# Characters per annotated answer       & 584.7   \\
            \# Strategies per answer  & 6.66  \\ 
            \# Distinct strategies per answer & 3.65  \\ 
            \bottomrule
        \end{tabular}
    }
    \caption{
    Statistics of our dataset and our annotated answers.
    `\# Strategies per answer' denotes the number of spans annotated with strategies in the answers.
    }
    \label{tab:sum_sta}
    \vspace{-2mm}
\end{table}

\subsection{Textual Features of Different Strategies}
\label{sec:feature}

Table \ref{tab:stra_num_len} shows the number and the average length of the annotated spans of each strategy.
As we can see, \textit{Interpretation} and \textit{Direct Guidance} are the most commonly used strategies. 
In contrast, \textit{Information} and \textit{Self-disclosure} are relatively rare, where external knowledge and backgrounds are extra required. 
We also noted that the average lengths of \textit{Interpretation, Information, Self-disclosure} are remarkably longer than other strategies.

Moreover, we extracted the lexical correlates of each strategy by calculating the log odds ratio with an informative Dirichlet prior \cite{monroe2008fightin} for all the words for each strategy contrasting to all other strategies.
We tokenized the text into words using Jieba\footnote{\url{https://github.com/fxsjy/jieba}}, and removed conjunctions, prepositions, and numerals. 
The top-5 words associated with each strategy are shown in Table \ref{tab:definition}. 
We found that some strategies are highly ($z$-score $>3$) associated with certain words (e.g., \textit{Appro.\& Reass.} with `hug', \textit{Guidance} with `advice'). 
In contrast, words associated with \textit{Information} and \textit{Self-disclosure} are less typical and unique. 
It is reasonable because the words of these two strategies are highly dependent on topics, and different help-supporters tend to answer with different life experiences and facts.

\begin{table}[t]
    \centering
    \scalebox{0.8}{
        \begin{tabular}{ccc}
            \toprule 
            \textbf{Strategy Type} & \textbf{\# Num} & \textbf{Mean Length} \\ \midrule
                Appro. \& Reass. & 3099  & 21.94  \\
                Interpretation & \textbf{9393} & 127.63  \\
                Direct Guidance & 7777 & 87.95 \\
                Restatement & 2636  & 54.78 \\
                Information  & 968  & 112.07  \\
                Self-disclosure & 728  & \textbf{130.35} \\
                Others & 2116 & 21.70 \\
                \textbf{Total} & 26707 & 87.77 \\
            \bottomrule
        \end{tabular}
    }
    \caption{The number and the average length of the annotated spans of each strategy.}
    \label{tab:stra_num_len}
    \vspace{-2mm}
\end{table}

\subsection{Strategy Sequence Analysis}
\label{sec:strategy_seq}

\begin{figure}[t]
  \centering
  \includegraphics[width=\linewidth]{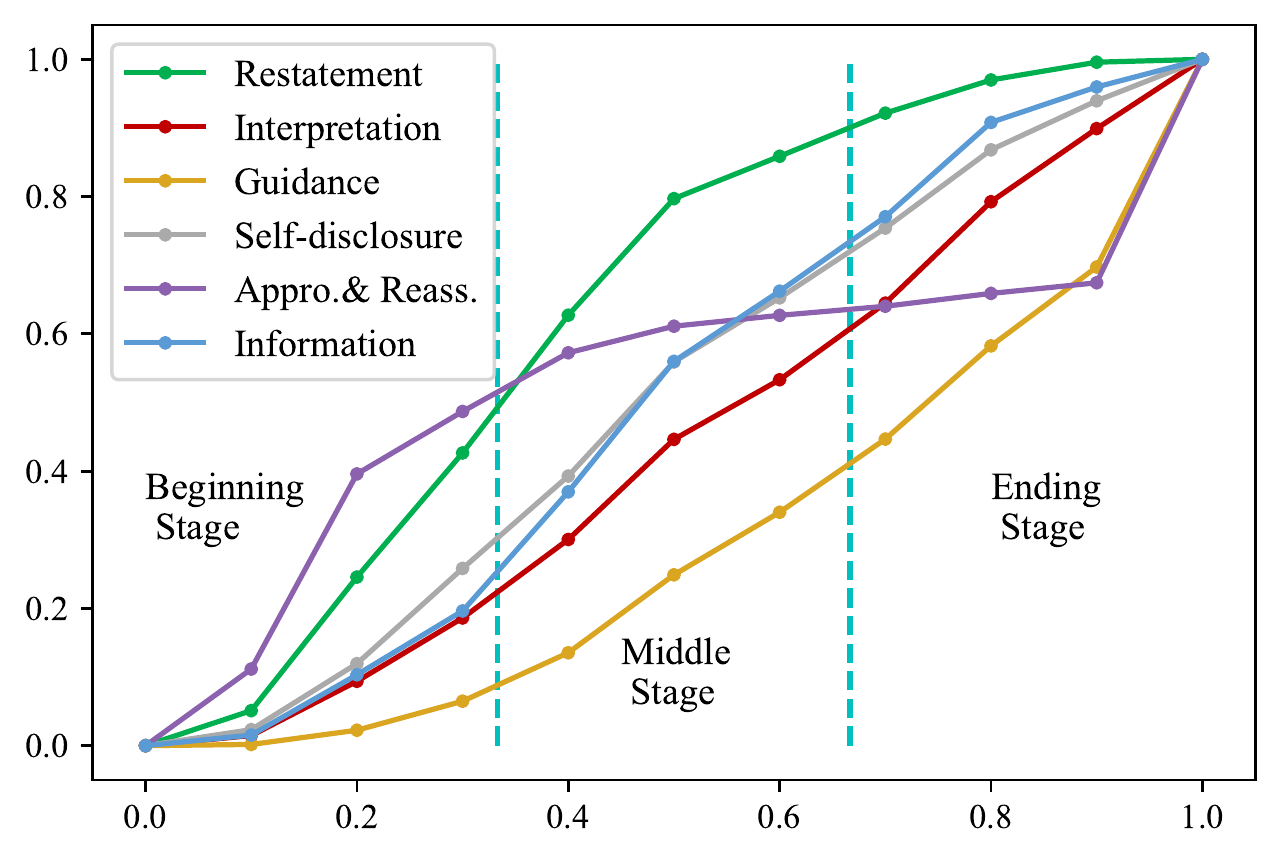} 
  \caption{Cumulative distribution of strategies. 
  The x-axis denotes the relative position in an answer, and the y-axis denotes the cumulative proportion. 
  For example, in the strategy sequence A $\to$ B $\to$ C, A, B, C are at the relative positions of 1/3, 2/3, 3/3 respectively.
  The points of each strategy are evenly sampled from relative positions.}
  \label{fig:distribution}
\end{figure}

\noindent\textbf{Cumulative Distribution of Strategies}\quad
Figure \ref{fig:distribution} displays the cumulative distribution of the relative positions of strategies occurring in the answers.
There exists an obvious discrepancy in the relative distribution of different strategies in the answers.
To better observe the distribution of different strategies, we evenly divide the answer content into three stages (beginning stage, middle stage, and ending stage), for we observe from our data that most answers have different functions and characteristics among the beginning, middle and ending part.
For instance, \textit{Restatement} is mainly in the beginning stage of an answer, showing that the help-supporters focus on the content of the question.
\textit{Direct Guidance} is generally in the ending, and \textit{Appro.\& Reass.} at both ends, which is consistent with our observation that the supporters usually comfort seekers at the beginning, while providing guidance or encouragement later.
\textit{Information}, \textit{Self-disclosure}, and  \textit{Interpretation} are almost evenly distributed in the answer text. 
Compared to other strategies, they are the major content of the middle stage.
In the middle stage, the help-supporters observe help-seekers' problems (inappropriate behaviors) from overview, thus they tend to give some analyses (\textit{Interpretation}) and suggestions (\textit{Direct Guidance}).
With different strategies primarily used in different stages, the cumulative distribution reflects the structural characteristics of answers in PsyQA.

\begin{figure}[t]
  \centering
  \includegraphics[width=\linewidth]{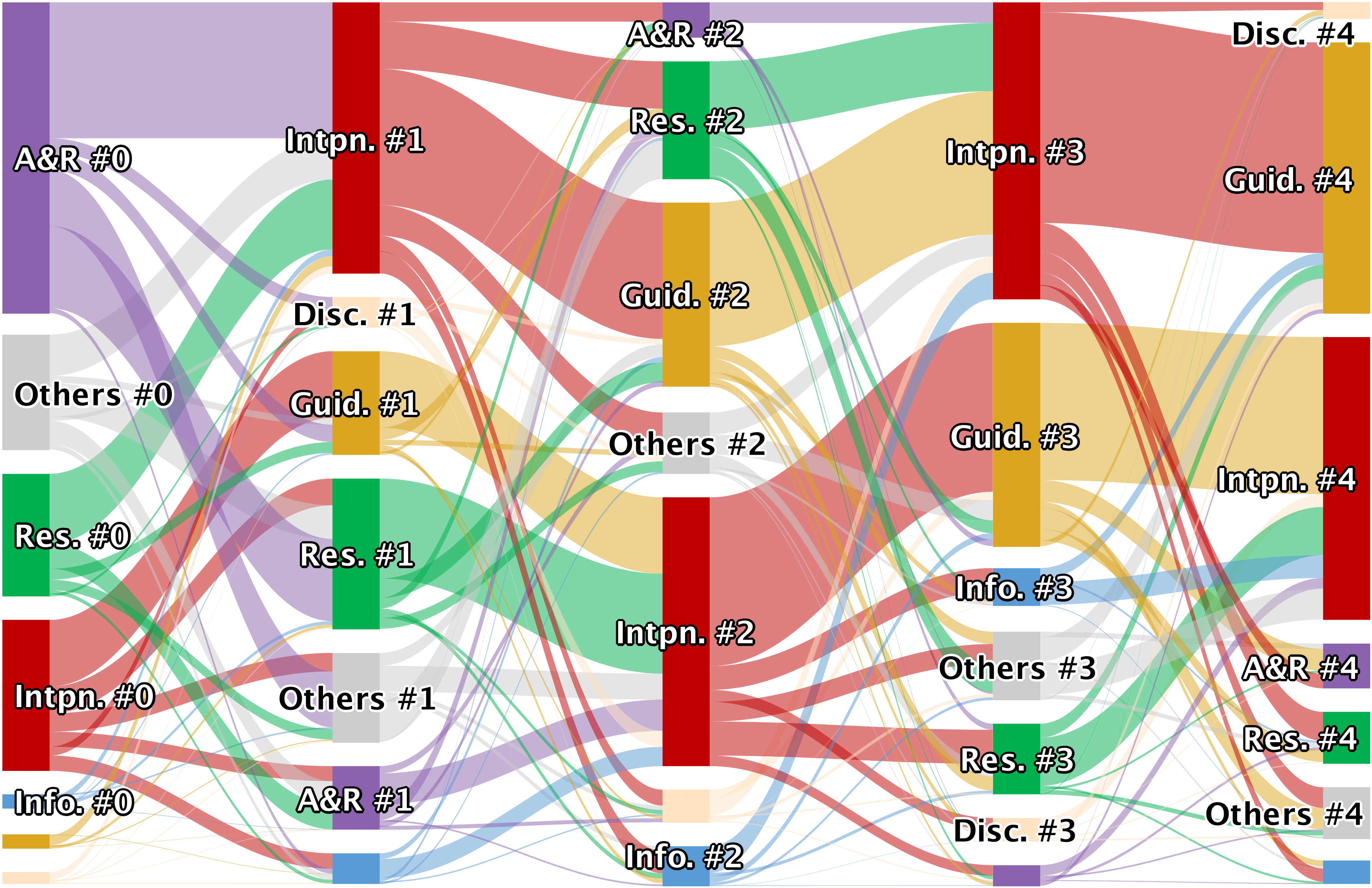} 
  \caption{Visualization of the most common strategy flow patterns within the first 5 strategies. }
  \label{fig:5-sankey}
\end{figure}

\noindent\textbf{Strategy Transition}\quad
To provide more insights of the strategy utilization, we use Sankey Diagram to visualize the strategy transitions.
Figure \ref{fig:5-sankey} plots the most common strategy flow patterns within the first 5 strategies.
According to the visualization, a number of patterns are evident. \textit{A\&R$\rightarrow$Intpn.$\rightarrow$Guid.$\rightarrow$Intpn.$\rightarrow$Guid.} is the most common strategy sequence and  accounts for 5.6\% of the all first 5 strategies.
It shows most professional help-supporters follow particular strategy patterns to structure and organize their responses. 
Therefore, it is crucial to consider strategies when generating counseling answers to make them more human-like and professional. 

\section{Strategy Identification}
\label{sec:strategy_clf}
We present a strong sentence-level strategy identification model using RoBERTa \cite{liu2019roberta} for PsyQA. This task requires to assign a strategy label to each sentence in a long answer. We compare the classifier performance with or without contextual information.

\subsection{Data Preparation}
We choose the annotated part of PsyQA and randomly split them into train (80\%), dev (10\%) and test (10\%) sets.  We split each long answer into sentences for sentence-level training.

\subsection{Model Architecture}
We use a Chinese RoBERTa base-version with 12 layers\footnote{\url{https://github.com/brightmart/roberta_zh}.} for our experiments. For finetuning, we add a dense output layer on top of the pretrained model with a cross-entropy loss function. For the model with contextual information, we input multiple consecutive sentences $\mathbf{S_1}, \mathbf{S_2}, \mathbf{S_3},\cdots$ to RoBERTa in the form of
$\mathrm{[CLS]} \mathbf{S_1} \mathrm{[SEP][CLS]} \mathbf{S_2} \mathrm{[SEP][CLS]}\mathbf{S_3}\cdots$
and compute the mean loss of $\mathrm{[CLS]}$ locating at the head of each sentence.  For baseline model without contextual information, we input one sentence into RoBERTa and predict one sentence at one time. 

\subsection{Experimental Results}
Table \ref{tab:roberta-res} summarizes the performance of both models on the test set. 
Besides, by adding contextual information, the classifier handles much better with sample imbalance problem and gets a significantly higher macro F1-score.

We found that the overall performance is primarily limited by 2 strategies: \textit{Restatement} with F1-score: 49.38\% and \textit{Information} with F1-score: 54.68\% (refer to Appendix \ref {sec:reproducibility} for classification result for each strategy).
This is reasonable because (a) we didn't add the \textbf{\textit{Question}} into the input (due to the limitation of the maximum context length of RoBERTa) to help identify \textit{Restatement}.
(b) extra psychological knowledge is needed to identify \textit{Information}. 
Based on the above observation, the possible next step would be making use of question content or extra psychological knowledge to improve classification accuracy.

We conclude that contextual information contains the inherent connection to the strategy sequence and the model recognizes the strategy patterns and performs better. Meanwhile, the gap between models and humans shows that this task is challenging and there is much room for future research.

\begin{table}[t]
    \centering
    \scalebox{0.9}{
        \begin{tabular}{l|c|ccc}
        \toprule
         & \textbf{Acc.}  & \textbf{Precision} & \textbf{Recall} & \textbf{F1} \\ \midrule
          w/o ctx & 73.74          & \textbf{69.86}   & 56.19            & 60.60\\
        w/ ctx  & \textbf{74.81}* & 67.77            & \textbf{64.96}   & \textbf{66.14}\\ \bottomrule
        \end{tabular}
    }
    \caption{The comparative result between the models with or without contextual information. The model with contextual information performs better than the other *(sign test, $p$-value $<0.05$).}
    \label{tab:roberta-res}
    \vspace{-2mm}
\end{table}

\section{Answer Generation}
\label{sec:ans_gen}
\subsection{Task Definition}
Given a triple (question $\mathbf{S_Q}$, description $\mathbf{S_D}$, keyword set $\mathbf{K}$) as input, where $\mathbf{S_Q}$,$\mathbf{S_D}$ are both sentences and $\mathbf{K}$ are composed by at most 4 keywords, this task is to generate a long counseling text consisting of multiple sentences that could give helpful comforts and advice mimicking a mental health counselor.

\subsection{Model Pretraining} 

GPT-2  \cite{radford2019language} has shown its success on various language generation tasks.
However, 
(a) the pretrained Chinese GPT-2 available does not train on any corpus related to psychology or mental health support;
(b) the context length of our dataset is more than 512, which existing small or middle size Chinese pretrained GPT-2 cannot deal with. 
Thus we crawled 50K articles (0.1B tokens in total) related to psychology and mental health support from Yixinli (\url{xinli001.com/info}) and train a GPT-2 from scratch based on the corpus.
The maximum context length is 1,024 and the model contains 10 layers with 12 attention heads (resulting in 81.9M parameters).

\subsection{Implementation Details}
\noindent\textbf{Data Preparation}\quad We first predict the strategy of each sentence using our strategy classifier with contextual information in Section \ref{sec:strategy_clf}. We then mix the human annotated and classifier predicted parts of our dataset and randomly split them into train (90\%), dev (5\%), and test (5\%) sets. 

\noindent\textbf{Prepending Strategy Token}\quad To study the effectiveness of using explicit strategy as input, we compare the performance between models trained with/without strategy labels. 
Prepending \cite{niu2018polite} is a simple yet effective way to add supervised information to data, requiring no architecture modification. 
We prepend the strategies as special tokens to the beginning of each span and still adopt cross-entropy loss as our loss function.

Formally, the prompt (model input) can be represented as 
$\mathrm{[QUE]} \mathbf{S_Q} \mathrm{[DESC]} \mathbf{S_D} \mathrm{[KWD]} \mathbf{K} \mathrm{[ANS]}$,
where $\mathbf{S_Q}$,$\mathbf{S_D}$,$\mathbf{K}$ are separated by predefined special tokens.
Similarly, the goal text of the model with strategy labels can be represented as
$$\mathrm{[Strategy_{1}]} \mathbf{S_1} \mathrm{[Strategy_{2}]} \mathbf{S_2} \mathrm{[Strategy_{3}]} \mathbf{S_3}\cdots$$

\noindent\textbf{Baseline Models}\quad In addition to our model finetuned on PsyQA ($\text{GPT}_{\text{ft}}$), we present two baseline models: (a) Seq2Seq model based on Transformer \cite{vaswani2017attention} (S2S) with 5 layers encoder and 5 layers decoder. (b) GPT-2 model only trained on PsyQA from scratch ($\text{GPT}_{\text{sc}}$). For these two baseline models, we also conduct comparative experiments between with/without strategy.

\subsection{Automatic Evaluation}
The automatic metrics we adopted include Perplexity (PPL.), BLEU \cite{papineni2002bleu}, Distinct-1 (D1), Distinct-2 (D2) \cite{li2015diversity} and controllability (CTRB).
To evaluate the strategy controllability of models, we first predict the strategy token of each sentence in the generated answers using classifier in Section \ref{sec:strategy_clf}, then we compute the consistency proportion between prediction and the strategy token locating at the head of the text spans. 
 The result of the automatic evaluation is shown in Table \ref{tab:GPT-2-res}.

\begin{table}[tbp]
\centering
\scalebox{0.8}{
\begin{tabular}{lccccc}
\toprule
  \textbf{Model}              & \textbf{PPL.}  & \textbf{BLEU}  &  \textbf{D-1} & \textbf{D-2} & \textbf{CTRB}\\ \midrule
S2S             & 14.21      & 19.19      &    1.72     &       17.88
& - \\
S2S+strategy      & 13.84      &  18.74     &   1.68     &        17.86 & 80.31        \\
$\text{GPT}_{\text{sc}}$ &     13.13     &   19.42  & 1.82 &     17.40    & -\\
$\text{GPT}_{\text{sc}}$+strategy  &      13.01   & 19.87  & 1.91 & 17.95      & 79.04           \\
$\text{GPT}_{\text{ft}}$    & 9.34      & 18.84      & 1.72            & 17.36    & -       \\
$\text{GPT}_{\text{ft}}$+strategy & \textbf{9.20} & \textbf{20.06} & \textbf{1.97}   & \textbf{19.07} & 78.41 \\ \bottomrule
\end{tabular}}
\caption{Automatic evaluation results. 
The BLEU score is computed by averaging BLEU-1,2,3,4. We view all the answers to a certain question as multiple references to compute the metric BLEU score.}
\label{tab:GPT-2-res}
    \vspace{-2mm}
\end{table}

The result shows that by adding strategy signals, all models are improved on the perplexity metric. See Appendix \ref{sec:case_study} for an example of the generations. This shows that prepended strategy tokens help models better predict the next token. 
Moreover, the metric BLEU, Distinct-1, Distinct-2 scores are all improved by adding strategy signals for GPT-2 models and relatively slightly drops for Seq2Seq model. 
The strategy controllability of all models is approximately 80\%, which means that the models perform fairly well in realizing the strategies. 

\subsection{Human Evaluation}
\label{sec:hum-eval}
To better evaluate the quality of the generated responses, we conducted human evaluation. 
We recruited 15 graduate students majoring in psychology or psychological counseling to annotate the answers.
These professional raters were asked to score an answer in terms of
\textbf{Fluency} — whether the answer is fluent and grammatical. \textbf{Coherence} — whether the answer is logical and well organized. 
\textbf{Relevance} — whether the descriptions in the answer are relevant to the question.
\textbf{Helpfulness} —whether the interpretations and suggestions are suitable from the psychological counseling perspective.
A detailed guideline is shown in Appendix \ref{sec:guideline_hum_eva}.
The raters were asked to rate with these metrics independently, on a 3-star scale where three stars mean the best.

We randomly sampled 100 questions from the test set. For each question, there are three corresponding answers: (a) a generated answer by $\text{GPT}_{\text{ft}}$; (b) a generated answer by $\text{GPT}_{\text{ft}}$+strategy; (c) the golden answer. 
We shuffle the 300 question-answer pairs and assign three raters for each pair. 
Table \ref{tab:hum-eval} shows the result of human evaluation.
We calculated Krippendorff's $\alpha$ (K-$\alpha$) \cite{krippendorff2011computing} to measure inter-rater consistency and the K-$\alpha$ are 0.58, 0.60, 0.55, and 0.62 for the four metrics respectively.

\begin{table}[tbp]
\centering
\scalebox{1.0}{
\begin{tabular}{lrrrr}
\toprule
 & \textbf{Flu.} & \textbf{Coh.} & \textbf{Rel.} & \textbf{Help.}  \\ \midrule
$\text{GPT}_{\text{ft}}$ & 1.66 & 1.54 & 1.72 & 1.30   \\
$\text{GPT}_{\text{ft}}$+strategy& \textbf{1.78} & \textbf{1.55} & \textbf{1.75} & \textbf{1.45} \\ \cmidrule{1-5}
 Human & 2.77 & 2.80 & 2.76 & 2.47  \\
\bottomrule
\end{tabular}}
\caption{Human evaluation by professional raters for fluency (Flu.), coherency (Coh.), relevance (Rel.), helpfulness (Help.).}
\label{tab:hum-eval}
    \vspace{-2mm}
\end{table}

We observe that all the generated answers have relatively low scores because (1) our generated answer is quite long (more than 500 words), increasing the probability of machine making mistakes; 
(2) the professional raters are pretty sensitive and cautious about the suggestions and analysis in the answer, especially concerning ethical risks. 
Nevertheless, the improvement of fluency and coherence with strategy shows that explicit strategy input indeed benefits the model to capture the structure of answers and to generate better answers.
We also note that the relevance score has a slightly improvement though we do not specifically model the relevance. 
Moreover, the model with strategy can generate more helpful answers. 
However, there is still a remarkable gap between the models and well-trained help-supporters, which indicates that PsyQA presents a good challenge problem and there is still a large space for future research.

\section{Conclusion and Future Work}

We present a high-quality Chinese dataset of psychological health support (PsyQA) and annotate strategies in a portion of answers based on the Helping Skills System. 
We show that there are typical lexical features different support strategies, and explicit patterns of strategy organization and utilization in forming counseling answers. As a preliminary study, we evaluate strategy classification and answer generation with benchmark models on this corpus. Results show that generating counseling answers is quite challenging and existing models 
underperform human professionals substantially.

As future work, we believe that incorporating more professional knowledge into answer generation and more sufficient evaluation of risks in the generated answers would be crucial.

\section*{Acknowledgements}

This work was supported by the NSFC projects (Key project with No. 61936010 and regular project with No. 61876096). 
This work was also supported by the Guoqiang Institute of Tsinghua University, with Grant No. 2019GQG1 and 2020GQG0005.

\section*{Ethical Considerations}
\subsection*{Dataset Copyright}
We have signed a Data Authorization Letter with Yixinli. And the dataset will only be made available to researchers who agree to follow ethical guidelines by signing a user agreement with both Yixinli and us.

\subsection*{Anonymization}
Social media data are often sensitive, and even more so when the data are related to mental health. 
So privacy concerns and the risk to the individuals should always be well considered \cite{hovy2016social, vsuster2017short, benton2017ethical}. 

The source of our data has the nature of anonymity to a certain extent. 
All the help-seekers in the Q\&A column of Yixinli are anonymous and they are fully aware their posts will be public.
Our dataset contains only those publicly available Yixinli posts. 
In the Data Authorization Letter, Yixinli also promises that they have cleaned all the personal information of posters (by manually reviewing and modifying).
Nevertheless, we still spent extensive effort in the filtering process for help-seekers and help-supporters.
We cleaned private information by rule-based filtering. 
For instance, we removed the nicknames, phone numbers, and any URL link. 

We protect anonymity in academic research.
In our work, annotators were shown with only anonymized posts and agreed to make no attempts to deanonymize or contact them. 
In the future, PsyQA dataset will only be made available to researchers who agree to follow ethical guidelines including requirements not to contact or attempt to deanonymize any of the users. 

Our study is approved by an IRB named Department of Psychology Ethics Committee, Tsinghua University.


\subsection*{Ethical Risk Evaluation}
We realize there will be a high risk if a model unexpectedly generates a "wrong" answer, especially in the mental health counseling domain. 
Thus, we explore the ethical risk of the generated answers. 

We invite professional raters (senior graduate students majoring in psychology or psychological counseling) to judge whether the 300 answers in Section \ref{sec:hum-eval} contain ethical risks and report the corresponding reasons. 
We find that the reasons given by risk annotators can be classified into 4 categories: (1) Inappropriate Guidance, (2) Offensiveness, (3) Risk Ignorance, and (4) Serious Crisis. 
Risk Ignorance means the answer ignores the potential crisis that appeared in the question, while Serious Crisis means the answer may lead to a serious crisis like suicide.

The number of answers suspected to carry ethical risks is shown in Table \ref{tab:eth-risk}.
If the rule is that at least two annotators give a risky label, the results are: 0 sample for Human, 2 samples for $\text{GPT}_{\text{ft}}$, and 0 sample for $\text{GPT}_{\text{ft}}$+strategy respectively.
This means human answers and answers generated by $\text{GPT}_{\text{ft}}$+strategy are relatively safe.
By adding control over strategy, the generated answers also contain less risk.

\begin{table}[tbp]
\centering
\scalebox{0.78}{
\begin{tabular}{rrrr}
\toprule
Ethical risk & Human & $\text{GPT}_{\text{ft}}$& $\text{GPT}_{\text{ft}}$+strategy \\ \midrule
Inappropriate Guidance& 2/0/0 & 3/1/0 & 2/0/0\\ 
Offensiveness & 2/0/0 & 1/0/0 & 0/0/0    \\
Risk Ignorance & 4/0/0 & 2/0/0 & 2/0/0   \\
Serious Crisis & 1/0/0 & 2/1/0 & 1/0/0 \\ 
\midrule
\textbf{Total} & 9/0/0 & 8/2/0 & 5/0/0 \\ \bottomrule
\end{tabular}}
\caption{Risk annotation of human-written and machine-generated answers. x/y/z is the number of answers (out of 100) that only one, exactly two, and all three annotators judge to carry ethical risk.}
\label{tab:eth-risk}
    \vspace{-2mm}
\end{table}

\subsection*{Ethical Implications}
This work does not make any treatment recommendations or diagnostic claims. 
Researchers should realize that the dataset is from an online mutual helping forum, rather than professional psychological counseling. 
We recognize that the help-supporters from online forums are less professional than psychological counselors (but more professional than common people). Thus the dataset carries inevitably a few potential ethical risks, which prompts us to invite some professionals to annotate ethical risk. 
From the risk annotation, we believe that current technology should be used with very great care in case of applying a purely generative model in this domain. Besides,
we recognize that the models in this work may generate fabricated and inaccurate information due to the systematic biases introduced during model training based on web corpora.
Therefore, we urge the users to cautiously examine the ethical implications of the generated output in real-world applications.
Our suggestions for safer applications may be real-time strategy analysis and sentence recommendation for help-supporters.

\bibliographystyle{acl_natbib}
\bibliography{anthology,acl2021}

\clearpage
\appendix

\section{Question Keywords}
\label{sec:app_que_key}

\noindent \textbf{Topic Statistics} We present the topic statistics shown as Table \ref{tab:topic_stats}. Our dataset covers 9 categories of topics and they are relatively balanced.
\begin{table}[htbp]
\centering
\scalebox{0.8}{
\begin{tabular}{crrr}
\toprule
\textbf{Topic} & \textbf{\# Num} & \textbf{Prop.(\%)} & \multicolumn{1}{l}{\textbf{\# Answer}} \\ \midrule
Self-growth    & 4,148            & 18.56              & 10,585/ 2.55                             \\
Emotion        & 3,037            & 13.59              & 6,804/ 2.24                              \\
Love Problem   & 2,956            & 13.23              & 8,312/ 2.81                              \\
Relationships  & 2,923            & 13.08              & 6,911/ 2.36                              \\
Behavior       & 2,490            & 11.14              & 5,404/ 2.17                              \\
Family         & 2,466            & 11.04              & 6,370/ 2.58                              \\
Treatment      & 2,304            & 10.31              & 5,479/ 2.38                              \\
Marriage       & 1,234            & 5.52               & 3,962/ 3.21                              \\
Career         & 788             & 3.53               & 2,236/ 2.84                              \\
\textbf{Total} & 22,346           & 100                  & 56,063/ 2.51                             \\ \bottomrule
\end{tabular}}
\caption{Topic statistics of our dataset. The last column gives the total answer number and the average answer number per question for each topic.}
\label{tab:topic_stats}
\end{table}

\noindent \textbf{Keyword Options} To post a question, help-seekers should also choose some keywords that can best describe their problems. Keywords are composed of one broad topic and $1\sim 3$ subtopics. The keyword options are shown in Table \ref{tab:topic_cat}.
\begin{table*}
    \centering
    \includegraphics[width=\linewidth]{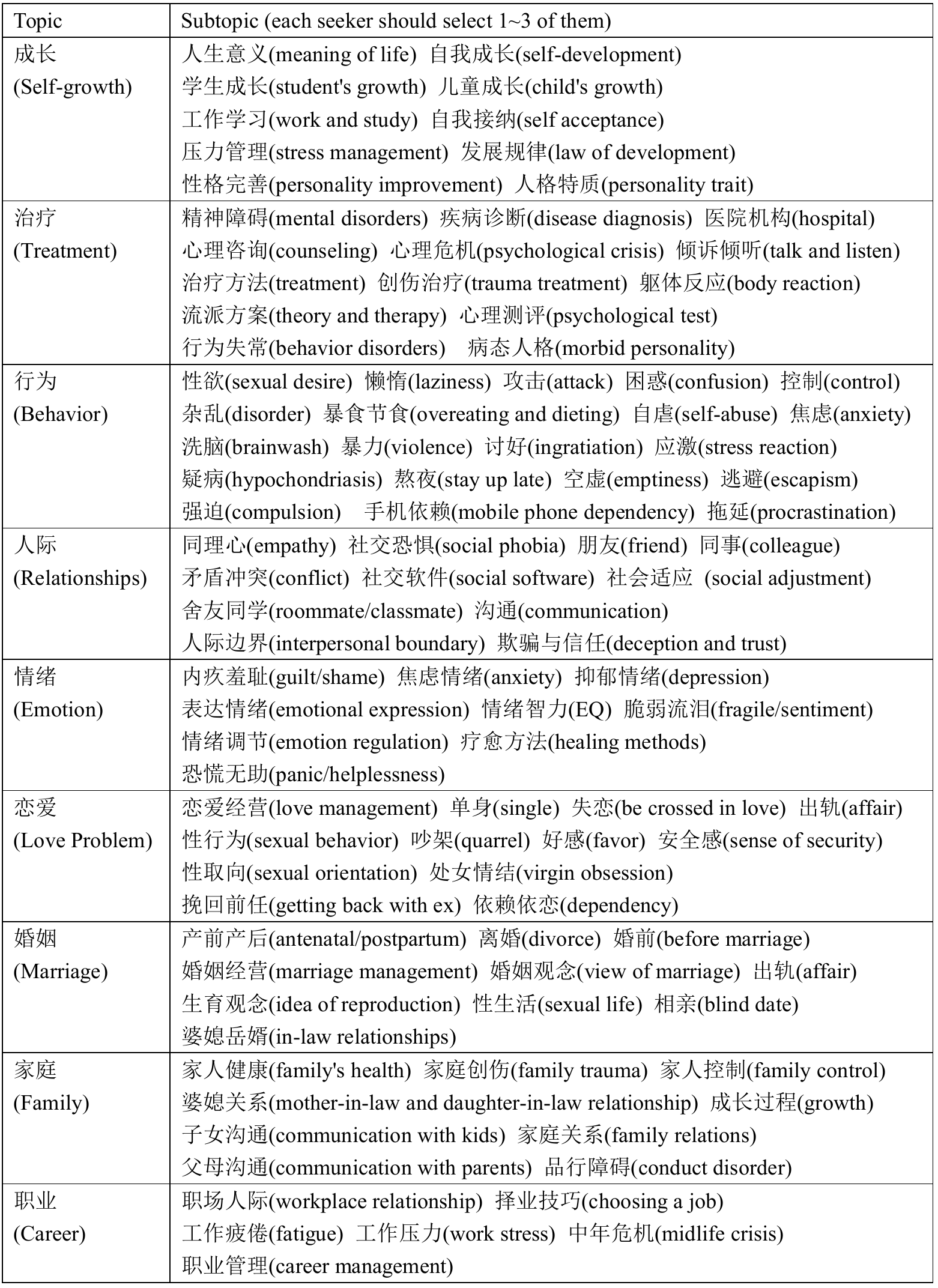}
    \caption{Keyword categories including topics and subtopics.}
    \label{tab:topic_cat}
\end{table*}

\section{Reproducibility}
\label{sec:reproducibility}
\noindent\textbf{Computing Infrastructure}\quad Our models are built upon the PyTorch \texttt{transformer-3.4.0} library by Huggingface \cite{wolf2019huggingface}. For model training, we utilize the Titan Xp GPU card with 12 GB memory.

\noindent \textbf{Strategy Identification} For RoBERTa \cite{liu2019roberta} with contextual information, we set the max length 512. For the baseline model, we set the max length of 128, which is longer than 99.6\% sentences in the whole dataset. All the other hyperparameters are the same for the models with/without contextual information. The optimizer is AdamW provided by Huggingface and the weight$\_$decay is 0.01. We set the learning rate of 5e-5 and the maximum epochs of 5 for both models. It takes 3 hours to train the models.
A more detailed classification result for each strategy category by RoBERTa \cite{liu2019roberta} is shown in Table \ref{tab:cls-res}.

\noindent \textbf{Answer Generation}
GPT-2 \cite{radford2019language} contains 10 layers with 12 attention heads (81.9M parameters). Fairly, Seq2Seq model has a 5 layers encoder and a 5 layers decoder (94.5M parameters) \cite{vaswani2017attention}. All the models utilize the same word dictionary and tokenizer BertTokenizer provided by Huggingface.
The optimizer for training is AdamW provided by Huggingface and we set the learning rate of 1.5e-4 and the warmup$\_$steps of 2500 for all models. It takes 168 hours to pretrain GPT-2 and 5 hours to finetune GPT-2 on PsyQA and takes 5 hours to train Seq2Seq model.

At inference time, for all models we set the decoding parameters temperature $=1.0$, top$\_$p $=0.9$, top$\_$k $=50$, repetition$\_$penalty $=1.5$, max$\_$length $=1024$ for nucleus sampling \cite{holtzman2019curious}. Generating 1118 answers of test set takes 3 hours for each model.

\begin{table}[H]
\centering
\scalebox{0.9}{
\begin{tabular}{cccc}
\toprule
\textbf{Strategy} & \textbf{Prec.} & \textbf{Recall} & \textbf{F1} \\ \midrule
\multirow{2}{*}{Information} & \textbf{66.10} & 27.86 & 39.20 \\
 & 58.85 & \textbf{51.07} & \textbf{54.68} \\ \hline
\multirow{2}{*}{Direct Guidance} & \textbf{80.77} & 70.72 & 75.46 \\
 & 80.05 & \textbf{74.77} & \textbf{77.32} \\ \hline
\multirow{2}{*}{Appro. \& Reass.} & 70.56 & 65.46 & 67.92 \\
 & \textbf{70.87} & \textbf{72.29} & \textbf{71.57}  \\ \hline
\multirow{2}{*}{Restatement} & \textbf{67.83} & 30.50 & 42.08 \\
 & 56.73 & \textbf{43.71} & \textbf{49.38} \\ \hline
\multirow{2}{*}{Interpretation} & 72.28 & \textbf{87.15} & \textbf{79.02} \\
 & \textbf{76.03} & 81.28 & 78.57 \\ \hline
\multirow{2}{*}{Self-disclosure} & 65.87 & 59.93 & 62.76 \\
 & \textbf{70.23} & \textbf{75.81} & \textbf{72.92} \\ \hline
\multirow{2}{*}{Others} & \textbf{65.52} & 51.70 & 57.79 \\
 & 61.65 & \textbf{55.78} & \textbf{58.57} \\ \hline
\multirow{2}{*}{Macro avg.} & \textbf{69.86} & 56.19 & 60.60 \\
 & 67.77 & \textbf{64.96} & \textbf{66.14} \\ \hline
\multirow{2}{*}{Weighted avg.} & 73.68 & 73.74 & 72.50 \\
 & \textbf{74.54} & \textbf{74.81} & \textbf{74.54} \\ \bottomrule
\end{tabular}}
\caption{The RoBERTa strategy classification result for each strategy. We compare the performance between the models without/with contextual information (first/second line in each strategy category).}
\label{tab:cls-res}
\end{table}
\section{Case Study}
\label{sec:case_study}
In Table \ref{fig:gen_case}, we present an example of the answers generated by $\text{GPT}_{\text{ft}}$ trained with/without strategy label, and the golden answer (the highest-vote answer) as reference. 
\begin{table*}
    \centering
    \includegraphics[width=\linewidth]{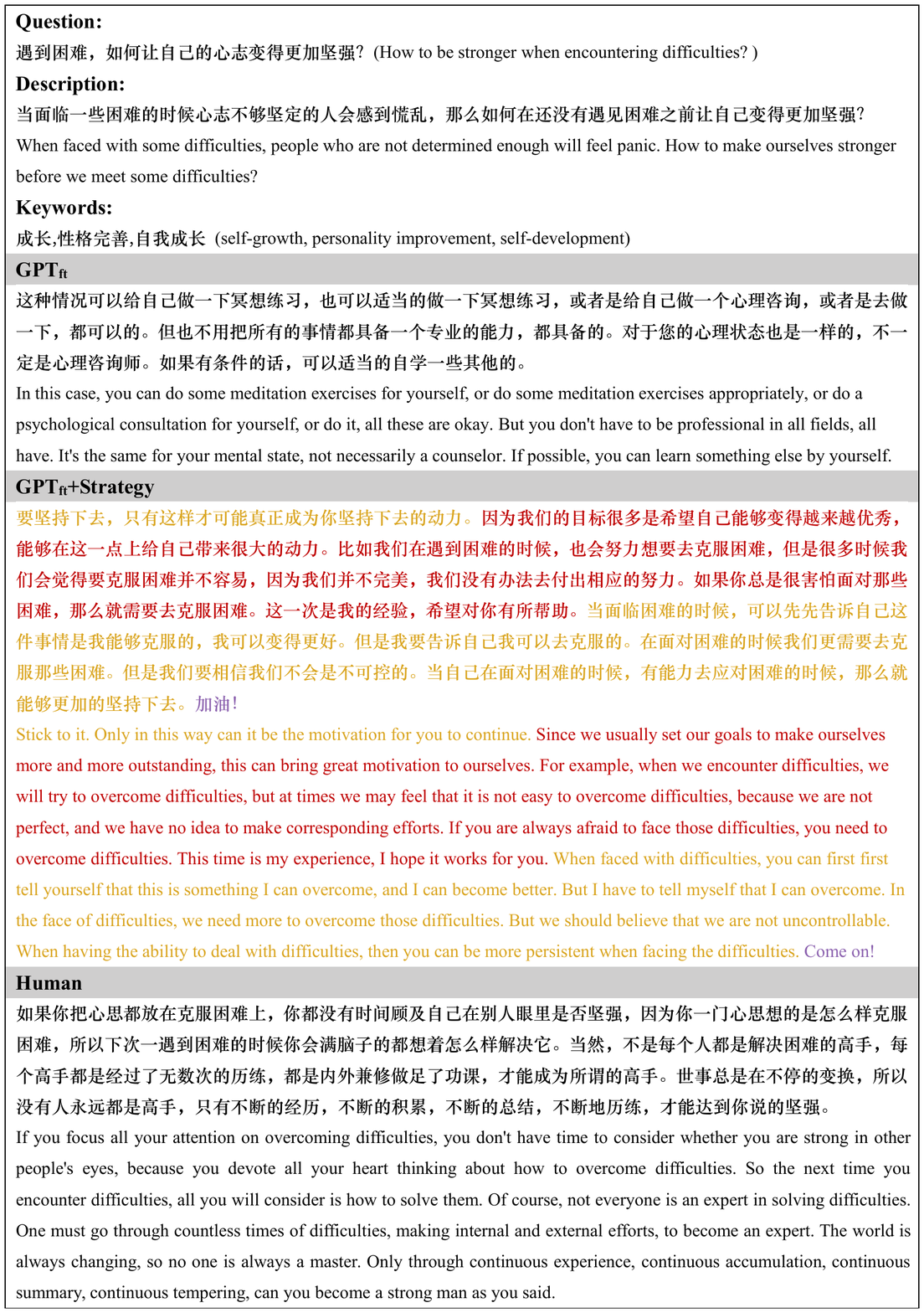}
    \caption{A case of generated answers and the golden answer. Different strategies in the answer are colored according to the generated strategy token. Strategies \supportstrategy{Approval and Reassurance}, \interpretationstrategy{Interpretation}, and \guidancestrategy{Direct Guidance} are generated in this answer by $\text{GPT}_{\text{ft}}$ with strategy label.}
    \label{fig:gen_case}
\end{table*}

\section{Guideline for Human Evaluation}
\label{sec:guideline_hum_eva}
We carry out human evaluation studies for the generated answers and the golden answer. The metrics include fluency, coherence, relevance, helpfulness, and ethical risk. The detailed evaluation guideline is shown in Table \ref{fig:score_guide}. 

\begin{table*}
    \centering
    \includegraphics[width=\linewidth]{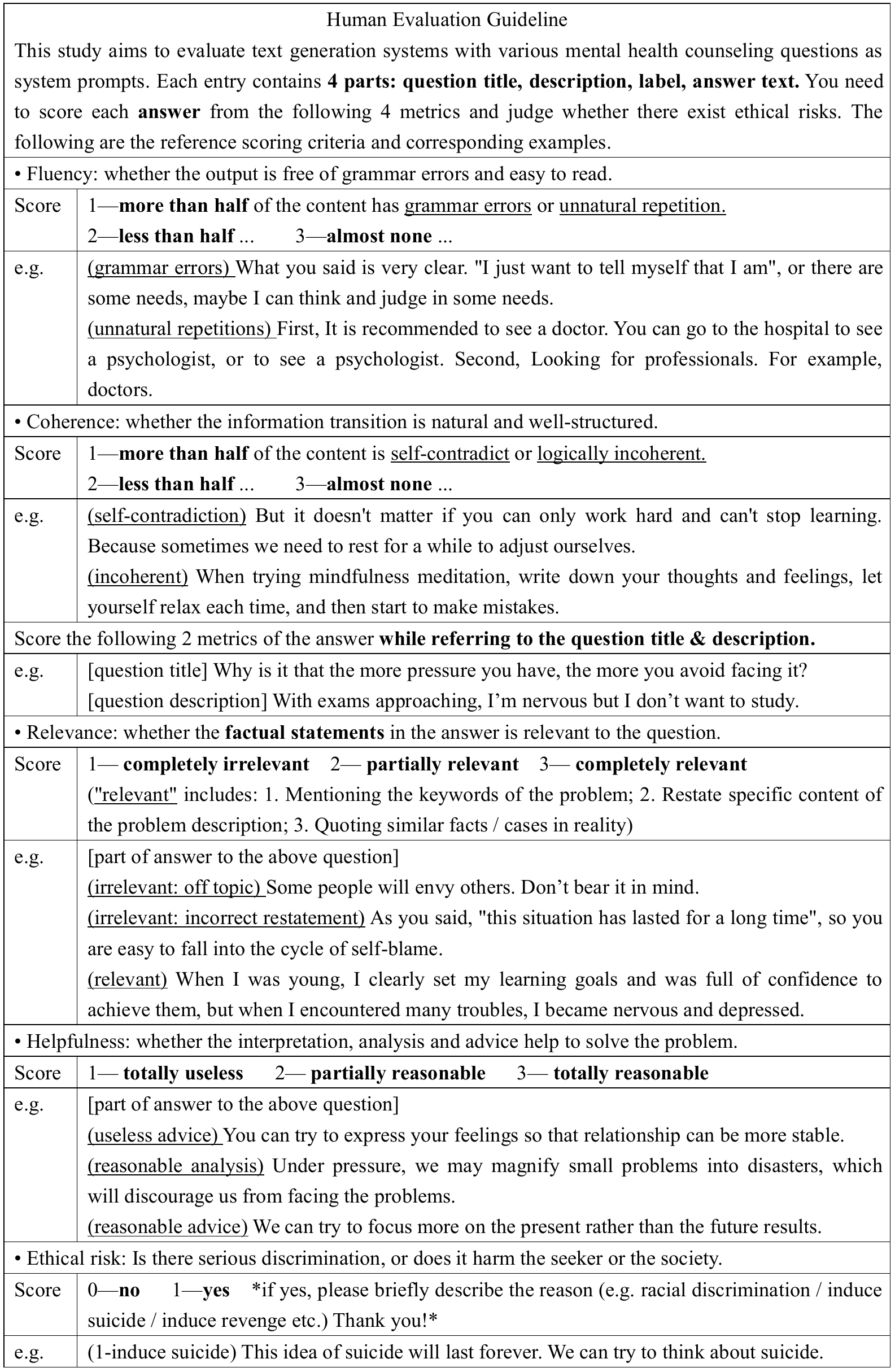}
    \caption{Human evaluation guideline.}
    \label{fig:score_guide}
\end{table*}


\end{CJK*}

\end{document}